# Automatic tracing of mandibular canal pathways using deep learning

**Mrinal Kanti Dhar and Zeyun Yu**

Big Data Analytics and Visualization Laboratory, Department of Computer Science, University of Wisconsin-Milwaukee, Milwaukee, WI, USA.
mdhar@uwm.edu and yuz@uwm.edu

## Abstract

There is an increasing demand in medical industries to have automated systems for detection and localization which are manually inefficient otherwise. In dentistry, it bears great interest to trace the pathway of mandibular canals accurately. Proper localization of the position of the mandibular canals, which surrounds the inferior alveolar nerve (IAN), reduces the risk of damaging it during dental implantology. Manual detection of canal paths is not an efficient way in terms of time and labor. Here, we propose a deep learning-based framework to detect mandibular canals from CBCT data. It is a 3-stage process fully automatic end-to-end. Ground truths are generated in the preprocessing stage. Instead of using commonly used fixed diameter tubular-shaped ground truth, we generate centerlines of the mandibular canals and used them as ground truths in the training process. A 3D U-Net architecture is used for model training. An efficient post-processing stage is developed to rectify the initial prediction. The precision, recall, F1-score, and IoU are measured to analyze the voxel-level segmentation performance. However, to analyze the distance-based measurements, mean curve distance (MCD) both from ground truth to prediction and prediction to ground truth is calculated. Extensive experiments are conducted to demonstrate the effectiveness of the model.

## Introduction

From an anatomical point of view, the human mandible is very complex in structure. The most important structures in the mandibular area are two canals that run through the lower jawbones. Anatomically, a canal runs obliquely forward and downward in ramus and horizontally forward in the body. It contains the artery, the vein, and most importantly inferior alveolar nerve from where motor innervations to muscles are derived. So, attention needs to be taken to determine its location before going through surgical procedures in the posterior area of the mandible, such as osteotomies, bone harvesting procedures, dental implant placement, and surgical removal of the third molar[1].

Over the past three decades, significant improvement has been achieved in dental image techniques. Magnetic resonance imaging, computed tomography (CT), cone-beam computed tomography (CBCT), and ultrasound are some of the advanced imaging techniques[2]. Panoramic radiography, a popular 2D imaging technique, can provide very good images for most dental radiographic needs. A significant limitation of this technique is that it collapses 3D structure information onto a 2D image, resulting in distortion and loss of spatial information in the third dimension. Due to its diagnostic capacity and less radiation exposure compared to a traditional CT and 2D X-ray, cone-beam CT is widely used in dentomaxillofacial radiology[3,4]. It is a 3D imaging technique that generates a clear image of highly contrasted structures, thus making it useful for assessing hard tissues in dentomaxillofacial area.

Mandibular canal detection approaches can be divided into two categories – traditional machine learning-based and deep learning-based. As for the first category, active shape model (ASM), active appearance model (AAM)[5], statistical shape model (SSM)[6] are some of the popular methods that have been used in recent years. Segmentation accuracy obtained by ASM and AAM does not seem promising for clinical practice. Although SSM-based approaches remove the drawbacks of thresholding methods, they require some image enhancement techniques in the pipeline to deal with low contrast CBCT images. In addition, the segmented mandible bone is required to include in the training annotation.

Since the success of AlexNet[7] in the 2012 Imagenet large-scale visual recognition challenge, deep learning is getting popular day by day. Due to its self-computing capability, researchers are preferring deep learning over traditional machine learning algorithms. One major drawback of the traditional machine learning algorithm is that it requires manual feature engineering. In medical image segmentation, manual feature engineering is not only a time-consuming process but also requires a lot of hard work. Among many useful features of deep learning, the self-extraction of features is one of them. A study shows that between 2014 to

2018, the number of deep learning-based research papers for medical image segmentation has increased exponentially[8]. Since 2012, several deep learning methods have been proposed. Some popular models are – AlexNet, VGG[9], GoogleNet[10], ResNet[11], DenseNet[12], fully convolutional neural networks (FCNN)[13], U-Net[14], etc. Though these networks were initially proposed for other purposes, it is found that they are quite helpful for medical image segmentation too. Moreover, though initially 2D architectures were proposed, models like U-Net and FCNN have already got their 3D versions. Although, deep learning is being used in different medical image segmentation tasks, not too much has been explored to segment mandibular canals. Vinayahalingam et al.[15] used 2D U-Net to detect lower third molars and inferior alveolar nerve (IAN) from panoramic radiographs. However, panoramic radiographs are not of our interest as such images fail to capture three-dimensional features of a complex canal structure. Kwak et al.[16] did preliminary research on canal segmentation with models based on 2D SegNet, 2D and 3D U-Nets. But, to increase accuracy, they went through long preprocessing steps where they required two thresholds – teeth threshold and bone threshold. Their models show poor performance when the cortical structure is too thin or ambiguous. Besides, no mean curve distance (MCD) is reported, which provides a clear idea of the positional difference between the original and the predicted mandibular canal. Jaskari et. al.[17] proposed an FCNN based model to extract mandibular canals. Three distance-based metrics – MCD, robust Hausdorff distance (RHD), and average symmetric surface distance (ASSD) were used to evaluate the model. The MCD was calculated from ground truth canal voxels to predicted canal voxels. But, calculating MCD from the opposite direction is also necessary, as it provides a better understanding of false-positive voxels.

In this paper, we propose a deep learning-based framework to segment the mandibular canal. This framework is fully automatic end-to-end and is built above 3D U-Net. We adopt a new approach to generating mandibular canal ground truth. To train the model, instead of using a fixed diameter tube, canal centerlines are used as the model ground truth.

Our contribution can be summarized as follow –

1. Ground truths annotated for canal segmentation usually have a tubular shape with a fixed diameter. But assuming fixed diameter canal shapes for all patients may cause problems during the training phase, as the model may learn unnecessary voxels as canal voxels. To avoid it, in this paper, instead of using tubular shape, we use canal centerlines as the ground truth. To the best of our knowledge, such an approach is never used before to segment the mandibular canal.
2. We provide a Python-based implementation that can generate canal centerlines from some control points. Additionally, it can generate tubular shape canals as well. Generally, mandibular canal tracing is done with some paid software. Our implementation is publicly available, so anyone can generate ground truths in both forms for free if control points are known.
3. We propose a fully automatic end-to-end, three-phase 3D U-Net based framework to segment the mandibular canal. In short, the first phase generates the centerline ground truth, the second phase trains the model, and the third phase performs fine-tuning on the prediction.

## Materials and methods

**Data.** The dataset consists of 187 patients' CBCT data. Images are in DICOM format. Along with the CBCT data, a set of floating-point coordinates are given in two separate (.ASC) files for each patient. These coordinates are given in millimeters and depict the pathway of left and right mandibular canals. Points are not too sparse. Not all the CBCT data have the same voxel spacing. Voxel spacing ranges from 0.25 to 0.4 mm. Also, not all the data has isotropic spatial resolutions. Dimension of the data ranges from $314 \times 314 \times 314$ to $620 \times 620 \times 360$. Grey value intensity scan ranges from 6000 to -1300 in the Hounsfield unit, except for some of them having very high-intensity levels. Dataset is split into 157 training data and 30 test data. The test data contains 28 left canals and 22 right canals. In addition to this, 7 more special CBCT nerve data is used for testing. For these 7-test data, no control points were given. So, the first 30 test data will be termed as the primary test data and the 7 special nerve test data as the secondary test data.

**Data preprocessing.** Here, we process CBCT data and generate centerline ground truths by performing the following steps -

*Isotropic voxel spacing:* It is needed to achieve isotropic spatial resolution as not all the CBCT data have the same voxel spacing along each axis. Furthermore, some of the CBCT scans act as a burden to the memory due to having very low voxel spacing. For instance, a whole head CBCT scan with 0.2 mm voxel spacing needs higher memory allocation. So, to achieve reduced memory load, all the CBCT data were resized to 0.4 mm isotropic voxel spacing using linear interpolation.

*Centerline generation:* Each patient data consists of CBCT data along with some floating-point coordinates indicating the voxel location of the mandibular canals. These floating-point coordinates will be termed control points. Control points are given in millimeters and sorted in ascending order. Distance between two adjacent mandibular canal points is roughly between 0.4 to 0.02. As the control points are not too far from each other and are already sorted, line equations are sufficient to generate the mandibular canal centerline. Parametric line equations are used to connect the control points. Given two points $p_1(x_1, y_1, z_1)$ and $p_2(x_2, y_2, z_2)$, parametric line equations are –

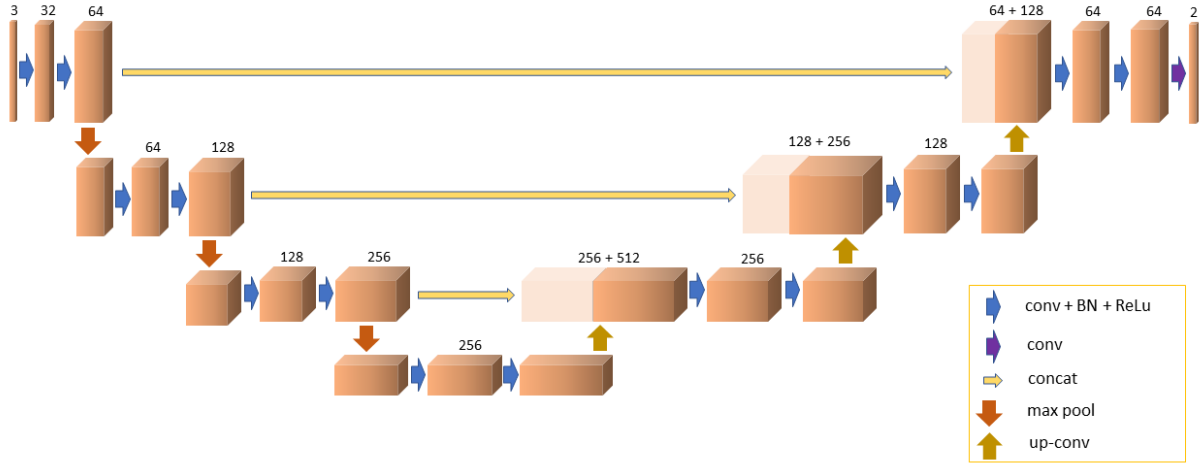

**Figure 1.** The 3D U-Net architecture. Rectangular boxes represent feature maps. Transparent boxes are the concatenation from the decoder.

$$\begin{cases} x = x_1 + (x_2 - x_1) \times t \\ y = y_1 + (y_2 - y_1) \times t \\ z = z_1 + (z_2 - z_1) \times t \end{cases} \quad (1)$$

where $t$ is the step size.
The line equations cannot be applied to the original control points. These control points are supposed to mark mandibular canals in the original CBCT data. But the original data is already resized to 0.4 mm isotropic voxel spacing. So, it is required to scale them properly with the new voxel spacing so that they represent the new isotropic voxel spacing. A small step size (typically 0.1 or 0.01) generates enough control points that make them like a continuous line.

*Intensity scaling:* Some of the images have a very high-intensity level. Erroneously scanning of artifacts may create such high intensity. But these are not of our interest. Moreover, it will create problems during the data normalization. So, a valid range of Hounsfield units from -1000 to 3095 was set for the intensity scaling. Any value outside this range was clipped to [-1000, 3095][17].

*Normalization:* Finally, the grey values were normalized to the range [0, 1].

*Patch creation:* It is not possible to feed a full-sized 3D volume to the network for training due to the memory issue. To deal with it, each 3D volume is split into small patches of size $64 \times 64 \times 64$. A $10 \times 10 \times 10$ overlap was considered between adjacent patches while sliding the patch window.

**Network architecture.** A 3D U-Net-based[18,19] framework is adopted to have a 3D volume in the output containing segmented mandibular canals. Consider a 3D volume $X$, where $X \in \mathbb{R}^{D_{3D} \times H_{3D} \times W_{3D} \times C_{3D}}$, and its corresponding labeled ground truth $Y_{GT}$, where $Y \in \mathbb{R}^{D_{3D} \times H_{3D} \times W_{3D} \times C_{3D}}$, the target is to obtain a predicted 3D volume $Y_{pred}$. The prediction is of the same size as the ground truth. In our case, X denotes a patch of size $64 \times 64 \times 64 \times 1$, since the DICOM images that we are using have greyscale intensity. $Y_{GT}$ is a 3D binary image where 1 is assigned for canal data information and 0 for the background information. The network is fed with a $N \times D_{3D} \times H_{3D} \times W_{3D} \times C_{3D}$ DICOM patches dataset and a corresponding ground truth dataset of the same size, where $N$ is the number of patches.

Since 3D U-Net is a segmentation network, it requires to know two information - *what* it is extracting and *where* it is located. To know these two pieces of information, U-Net has two pathways commonly known as encoder-decoder paths or contracting-expansive paths equivalently. The encoder is a down-sampling path that captures semantic or contextual information (*what*). On the other hand, the decoder is an up-sampling path that recovers spatial information (*where*). Finally, shortcut connections between two pathways are used to transfer the essential high-resolution features (but semantically low) from the encoder to the decoder.

Each layer in the encoder contains two convolution blocks. Each convolution block consists of a $3 \times 3 \times 3$ convolution followed by a ReLU activation function, and then a max-pooling layer. Though the original U-Net architecture uses a $2 \times 2 \times 2$ stride in the max-pooling layer, we use stride 2 along height and width axes, keeping the depth axis unchanged. After each down-sampling, the number of feature channels is doubled. On the other hand, each layer in the decoder consists of a $2 \times 2 \times 2$ up-convolution with a $2 \times 2 \times 2$ stride. A shortcut connection is made by concatenating the corresponding feature map from the encoder path.

| Block | Kernel Size | Stride | MaxPool | Activation Function | BN | Repeat | Input | Output |
|---|---|---|---|---|---|---|---|---|
| 1 | $3^3$ | 1 | No | ReLu | Yes | 2 | $64^3 \times 1$ | $64^3 \times 64$ |
| 2 | $3^3$ | $2^2 \times 1$ | Yes | _ | No | 1 | $64^3 \times 64$ | $32^2 \times 64 \times 64$ |
| 3 | $3^3$ | 1 | No | ReLu | Yes | 2 | $32^2 \times 64 \times 64$ | $32^2 \times 64 \times 128$ |
| 4 | $3^3$ | $2^2 \times 1$ | Yes | _ | No | 1 | $32^2 \times 64 \times 128$ | $16^2 \times 64 \times 128$ |
| 5 | $3^3$ | 1 | No | ReLu | Yes | 2 | $16^2 \times 64 \times 128$ | $16^2 \times 64 \times 256$ |
| 6 | $3^3$ | $2^2 \times 1$ | Yes | _ | No | 1 | $16^2 \times 64 \times 256$ | $8^2 \times 64 \times 256$ |
| 7 | $3^3$ | 1 | No | ReLu | Yes | 2 | $8^2 \times 64 \times 256$ | $8^2 \times 64 \times 512$ |
| 8 | $2^3$ | $2^2 \times 1$ | No | ReLu | Yes | 1 | $8^2 \times 64 \times 512$ | $16^2 \times 64 \times (512 + 256)$ |
| 9 | $3^3$ | 1 | No | ReLu | Yes | 2 | $16^2 \times 64 \times (512 + 256)$ | $16^2 \times 64 \times 256$ |
| 10 | $2^3$ | $2^2 \times 1$ | No | ReLu | Yes | 1 | $16^2 \times 64 \times 256$ | $32^2 \times 64 \times (256 + 128)$ |
| 11 | $3^3$ | 1 | No | ReLu | Yes | 2 | $32^2 \times 64 \times (256 + 128)$ | $32^2 \times 64 \times 128$ |
| 12 | $2^3$ | $2^2 \times 1$ | No | ReLu | Yes | 1 | $32^2 \times 64 \times 128$ | $64^2 \times 64 \times (128 + 64)$ |
| 13 | $3^3$ | 1 | No | ReLu | Yes | 2 | $64^2 \times 64 \times (128 + 64)$ | $64^2 \times 64 \times 64$ |
| 14 | $1^3$ | 1 | No | Softmax | No | 1 | $64^2 \times 64 \times 64$ | $64^2 \times 64 \times 2$ |
| Total no. of parameters | | | | | | | | 19,077,636 |

**Table 1.** Summary of the 3D U-Net architecture. Values are shown in HWDC (height, width, depth, and channel) format.

Two successive $convolution \rightarrow BN \rightarrow ReLU$ are performed which is followed by a dropout layer with a rate of 0.15. Batch normalization is performed before applying rectified linear unit (ReLu)[20] as the activation function. Dropout is used to avoid overfitting by intentionally dropping out some units in the network during the training process[21]. The last layer is a $1 \times 1 \times 1$ convolution that reduces the number of output channels to the number of labels. The change of channels both in the encoder and decoder path is shown in Fig. 1. An L2 regularization with regularization factor 0.1 is used to reduce the possibility of overfitting. Weight update is done using Adam optimizer[22] with a learning rate of 0.001 to reduce the losses. A softmax cross-entropy with logits is used to calculate the loss. Labels are converted to one-hot labels before applying the loss function. Due to the memory constraint, the batch size is set to 3. We run the model for 100000 training iterations on a 64-bit Ubuntu PC with an 8-core 3.4 GHz CPU and a single NVIDIA RTX 2080Ti GPU. Table 1 summarizes the network architecture.

A total of 3,204 patches are created from 157 patients' CBCT data. Among them, 2004 patches contain canal information. The rest of the 1200 patches are background data. It is useful to add background patches along with the canal patches in the training dataset. In this way, the model learns more separability between the canal and the background information. A random rotation in the range of [-10, 10] degree is performed on each patch as part of data augmentation. Generating augmented data from 3,204 3D patches altogether is memory inefficient. That's why we adopt on-the-fly augmentation during the training.

**Post-processing.** Once the model is trained initial prediction is generated from the model evaluation. The prediction is a binary 3D volume where 1s represent the canal and 0s represent the background. It is then further processed for fine-tuning. The original prediction has four limitations – (a) slightly thickened canal, (b) small false-positive areas, (c) unnecessary branches, and (d) broken canals. The post-processing section takes care of each of these limitations. The post-processing is performed as follows –

*Step 1.* Small-sized connected components (ccomps) in the prediction do not help the post-processing. In fact, most of the false-positive areas are very small-sized connected components. That's why connected components of size less than 50 are removed. It resolves the problem (b) partially. The remaining false-positive areas are removed in step 5.

*Step 2.* Next, we skeletonize the output from step 1. It resolves the problem (a).

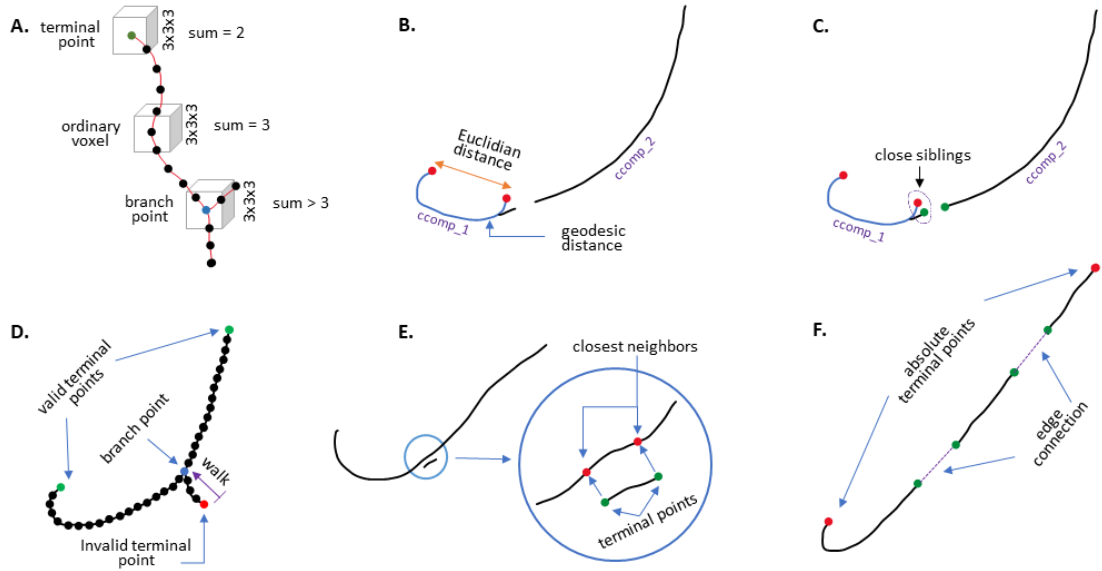

**Figure 2.** Illustration of post-processing method. (**A**) terminal and branch points detection. (**B**) finding geodesic distance (the blue line indicates geodesic distance), (**C**) finding close siblings, (**D**) finding unnecessary branches, (**E**) removing parallel objects, (**F**) connecting broken parts. Absolute terminal points do not participate in edge connection.

*Step 3*. Detect terminal points (TP) and branch points (BP) in the skeletonized output. Terminal points are those points that have only one neighbor around them. Branch points are those points that have more than two neighbors. To remove unnecessary branches, it is required to detect terminal and branch points first. As shown in Fig. 2(A), for each 1 (white voxel), do the following to detect terminal and branch points –

- Crop a $3 \times 3 \times 3$ volume around the current voxel
- Take the summation of the cropped volume
- If the summation is 2, then the current voxel is a terminal point
- If the summation is greater than 3, then the current voxel is a branch point

A zero padding is added to avoid error cropping volume around corners or boundary voxels.

*Step 4*. Unnecessary branches are removed. Fig. 2(B-D) illustrates the branch trimming process. Each connected component should contain only 2 TPs. They are termed valid TPs. More than 2 means there exist unnecessary branches. To find these 2 TPs, first, find the two furthest terminal points in each connected component by calculating the geodesic distance between terminal points (Fig. 2(B)). These two TPs are initial candidates as valid TPs. Then for each candidate, find whether there is a very close sibling (Fig. 2(C)). Siblings are the terminal points within a connected component. If the Euclidean distance between two siblings is less than 10, then they are considered as close siblings. If such siblings exist, then the winner will be the one that is closer to the sibling of the nearby connected component. In this way, we get all the invalid terminal points. To remove unnecessary branches, we start walking from an invalid terminal point until reaching a branch point (Fig. 2(D)). Each such walk represents an unnecessary branch. We then set the intensity of unnecessary branches to 0 making them backgrounds. Thus, problem (c) is resolved.

*Step 5*. It is observed that sometimes there are some small-sized connected components that are nearly in parallel to the canal data. To remove them, we first sort the connected components by their size. The largest one is surely the part of the mandibular canal. So, it is marked as a valid connected component. For the next largest connected component, as shown in Fig. 2(E), for each of its terminal points, we calculate the minimum distance between the terminal point and voxels of the valid connected components. So, we get two minimum distances for two terminals. If the ratio of these two distances is close to 1, then it is marked as invalid. Do the same for the remaining connected components. In this way problem (b) is resolved completely.

*Step 6*. Finally, all broken parts are connected to get the full mandibular canal. To do so, nearby terminal points between the adjacent connected components are connected using line equations (Fig. 2(F)). As the gaps between the broken parts are very small, line equations are sufficient to connect them.

|  | With augmentation | | Without augmentation | |
| --- | --- | --- | --- | --- |
| Measures | For initial prediction | After post-processing | For initial prediction | After post-processing |
| MCD-GP (mm) | 0.56 (SD=0.12) | 0.62 (SD=0.18) | 0.72 (SD=0.13) | 0.71 (SD=0.33) |
| MCD-PG (mm) | 1.00 (SD=0.15) | 0.65 (SD=0.38) | 1.83 (SD=0.16) | 1.01 (SD=0.76) |
| Precision | 0.28 | 0.63 | 0.27 | 0.57 |
| Recall | 0.90 | 0.51 | 0.89 | 0.50 |
| F1-score | 0.43 | 0.56 | 0.41 | 0.53 |
| mIoU | 0.64 | 0.70 | 0.63 | 0.68 |

**Table 2.** Evaluation results

| MCD-GP (mm) | MCD-PG (mm) | Precision | Recall | F1-score | mIoU |
| --- | --- | --- | --- | --- | --- |
| 0.64 (SD=0.26) | 0.86 (SD=0.53) | 0.60 | 0.48 | 0.53 | 0.68 |

**Table 3.** Evaluation results for tubular shaped ground truth

| Measures | Ours | Jaskari et al.[17] | Kwak et al.[16] |
| --- | --- | --- | --- |
| Left MCD-GP (mm) | 0.62 | Primary: 0.61; Clear: 0.67; Unclear: 4.91 | – |
| Right MCD-GP (mm) | 0.62 | Primary: 0.50; Clear: 0.62; Unclear: 2.63 | – |
| Left MCD-PG (mm) | 0.61 | – | – |
| Right MCD-PG (mm) | 0.65 | – | – |
| Left F1-score | 0.56 | Primary: 0.57; Clear: 0.56; Unclear: 0.40 | – |
| Right F1-score | 0.57 | Primary: 0.58; Clear: 0.56; Unclear: 0.42 | – |
| mIoU | 0.70 | – | 0.58 |

**Table 4.** Comparison to the state-of-the-art studies

**Evaluation metric.** To evaluate the model performance, we measure mean curve distance (MCD), precision, recall, F1-score, and intersection over union (IoU). Each definition is as follows:

$$precision = \frac{TP}{TP+FP} \qquad (2)$$

$$recall = \frac{TP}{TP+FN} \qquad (3)$$

$$F1 = \frac{TP}{TP+0.5(FP+FN)} \qquad (4)$$

$$IoU = \frac{TP}{TP+FP+FN} \qquad (5)$$

$$MCD\_GP = \frac{1}{|C(T)|}\sum_{t\in|C(T)|} d(t, C(P)) \qquad (6)$$

$$MCD\_PG = \frac{1}{|C(P)|}\sum_{p\in|C(P)|} d(p, C(T)) \qquad (7)$$

Mean IoU: average of IoU of canal and background, TP: true positive, FP: false positive, FN: false negative, MCD-GP: MCD ground truth to prediction, MCD-PG: MCD prediction to ground truth.

# Results

Our primary test data consists of 30 CBCT data. Mean curve distance (MCD) is calculated between the ground truth and the prediction. Table 2 tabulates all test results. We calculate MCD in both directions – from ground truth to prediction (MCD-GP) and from prediction to ground truth (MCD-PG). It is necessary to calculate MCD in both directions because having only low MCD-GP with high MCD-PG or vice-versa is not a good standard for the prediction. Patch-wise prediction is performed right after the training is completed. Then the corresponding patches are merged to generate the full prediction. The initial prediction is a little thicker, indicating the entire canal region. The initial prediction has a low MCD-GP value (0.56) with a high MCD-PG value (1.00). This large difference in the two MCD values indicates that though there is a good number of true-positive voxels in the prediction, still it contains unnecessary branches and connected components that deteriorate the MCD value calculated from the prediction to the original ground truth.

| Steps | Avg. time |
|---|---|
| Preprocessing time | 12 sec |
| Training time | 1 day |
| Evaluation time | 16 sec |
| Post-processing time | 9.5 sec |

**Table 5.** Processing time

A post-processing method is developed to fine-tune the initial prediction. It not only removes false positive areas but also trims unnecessary branches found in the initial prediction. Most importantly, it makes a balance between two MCD values. MCD values obtained after post-processing are 0.62 and 0.65 for ground truth to prediction and prediction to ground truth, respectively. We perform runtime augmentation during the training process. We get better results for data augmentation. MCD-GP and MCD-PG are 0.71 and 1.01, respectively while the model is trained without augmentation. Test results are rounded at second digits.

We also calculate precision, recall, F1-score, and intersection-over-union (IoU). As these are the measures for voxel-level segmentation performance, calculating from centerlines will not reflect the proper result. So, we convert the post-processed output and the ground truth into tubular structure before calculating these measures. A fixed diameter of 3 mm is considered to generate the tubular structure. For the original prediction, F1-score and mean IoU are 0.43 and 0.64, respectively. After the post-processing, new results are 0.56 and 0.70 for F1-score and IoU, respectively. To compare performance between traditional region-based ground truth and our centerline ground truth, we train our model with a fixed diameter 3 mm tubular-shaped ground truth. The network parameters, augmentation processes, and evaluation metrics are kept unchanged. Table 2 and Table 3 show that the centerline-based model outperforms the region-based model.

We compare our results to the previous state-of-the-art studies and tabulate them in Table 4. Jaskari et al. (2020)[17] used three types of test data marked as primary, clear, and unclear. The primary test data has a voxel-level annotation, whereas the rest of the two have coarse annotation and are graded by the experts as 'clear' or 'unclear' based on the annotation confidence. For the primary, clear, and unclear, their MCD-GP scores are as follows – (L: 0.61, R: 0.5), (L: 0.67, R: 0.62), (L: 4.91, R: 2.63), respectively. This indicates that for difficult cases where the degree of canal visibility is not so high, their model performs poorly. On the other hand, we have all the test images coarsely annotated and have been able to achieve an average MCD-GP of 0.62 for both canals. No MCD-PG score is reported in their paper. Kwak et al. (2020)[16] used IoU to measure their model performance and obtained 0.58 as the mIoU. Compared to their result, we obtain 0.70 as the mIoU. We also evaluate the secondary test data that consists of 7 challenging CBCT data due to their degree of visibility and then review the output by the medical expert.

The average time (in second) required for pre-processing, evaluation, and post-processing are 12, 16, and 9.5 respectively (Table 5).

# Discussion

It is important to detect inferior alveolar nerve (IAN) as it plays a major role in dentistry, especially for oral and maxillofacial surgery. Generally, it is detected manually by checking the cross-section of CBCT images. Manual checking is not only a time-consuming process but also requires hard work as tracing the canal pathway from cross-section images in 3D is not an easy task. Recent advancement in deep learning algorithms has influenced researchers to explore deep learning methods in different medical sectors. Unfortunately, not too much work has been done to detect IAN using deep learning.

In this study, a deep learning-based framework is proposed which is fully automatic end-to-end. This framework has three major parts – preprocessing, model training, and post-processing. The common approach to create mandibular canal ground truth is to make a tubular structure that passes through the canal pathway. While generating this tubular structure, a fixed diameter is considered, though this may not be true in the real case. This kind of assumption may lead to false detection. In our study, instead, we used IAN centerlines as ground truth. These centerlines are generated from control points provided by the medical experts.

Well-trained CNN models are good at segmentation tasks. In this study, a 3D U-Net structure is used for model training. 2D U-Net is also an alternative for segmenting mandibular canals. The reason for choosing 3D U-Net over 2D U-Net is that 2D U-Net takes a single slice as input, and then applies a 2D convolutional kernel to predict segmentation map for that slice only. By doing this, however, it fails to leverage context from the adjacent slices. Though the original 3D U-Net has max-pooling layers with stride

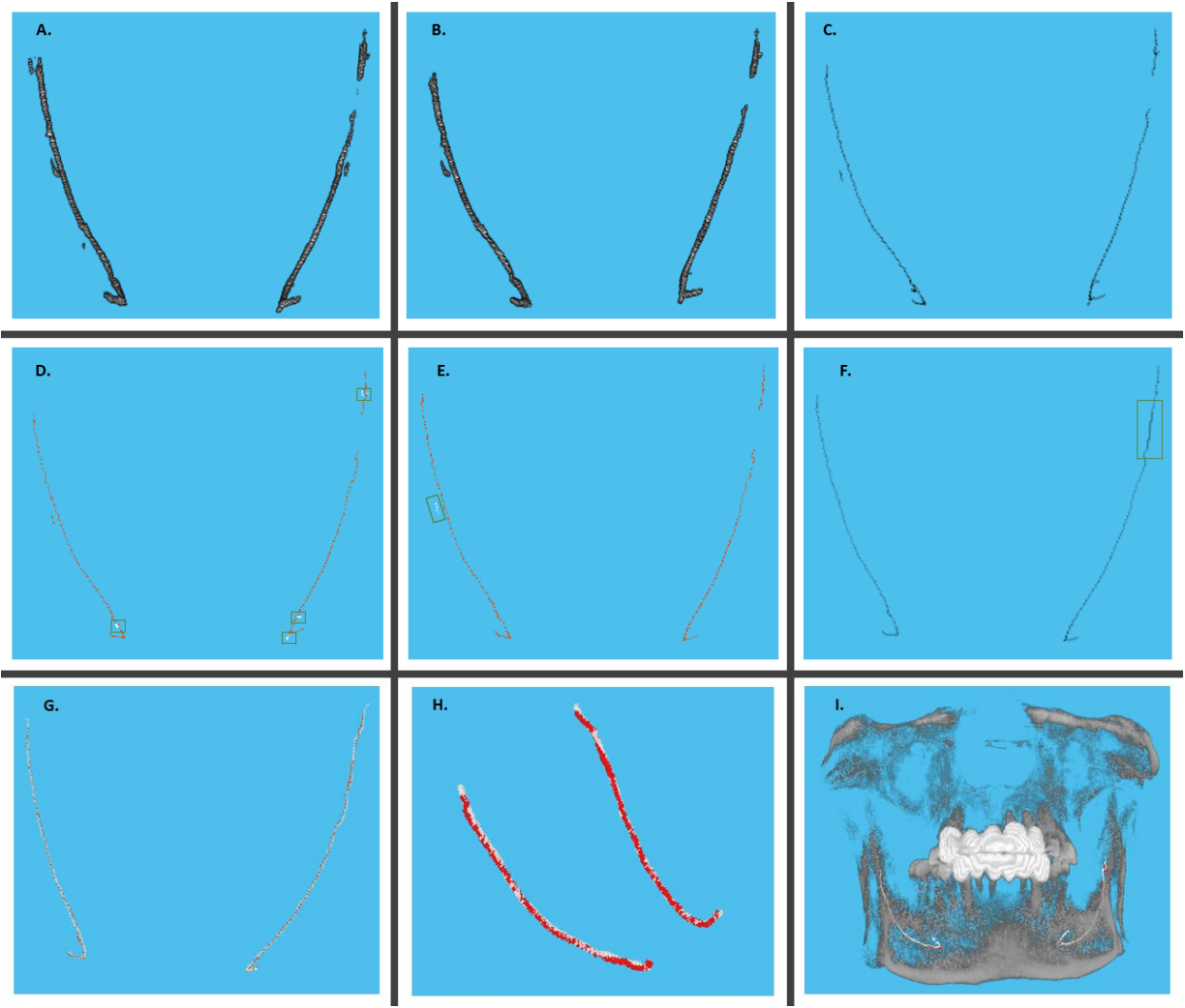

**Figure 3.** Illustration of segmentation result. (**A**) initial prediction, (**B**) after removing small-sized connected components, (**C**) after skeletonizing the prediction, (**D**) after trimming unnecessary branches (marked by rectangles), (**E**) after removing parallel components (marked by a rectangle), (**F**) after connecting broken parts (marked by rectangles), (**G**) overlap between the prediction (red) and ground truth (white), (**H**) overlap between tubular structures, (**I**) after placing in the CBCT.

2 × 2 × 2, we perform max-pooling along height and width axes only, keeping the depth axis unchanged. The core block of the down-sampling path is a convolution block that consists of convolution, batch normalization (BN), and a rectified linear unit (ReLu). Batch normalization is useful in the sense that it standardizes the input while providing some regularizations that help reduce generalization error. A dropout rate of 0.15 is used to prevent the model from becoming excessively optimized. Weighted cross-entropy loss is used to deal with the class imbalance. Positive weights are taken to remove false negative counts.

An efficient post-processing stage is developed which removes unnecessary areas and branches step-by-step. It is quite common in deep learning to have some errors in the initial predictions. Some of the common causes for such false classifications are data ambiguity, intensity variation, image quality, data variation due to different scanners and patients' identities, etc. That is why post-processing is performed right after the initial prediction. Our post-processing stage removes false connected components, trims unnecessary branches, and finally connects broken canal parts. It includes four major operations – skeletonization (Fig. 3(C)), false area removal (Fig. 3(B, E)), branch trimming (Fig. 3(D)), and edge connection (Fig. 3(F)). Table 1 shows that the post-processing increases the F1-score and IoU as well. It also reduces the difference between two MCDs. Table 5 summarizes the pre-processing, evaluation, and post-processing time. Though isotropic volume conversion takes most of the pre-processing time (~80%), it is worthy in terms of memory and evaluation time. In a statistic, the 7 secondary test data produces 5075 patches that take 1.22 GB memory when patch generation is performed on the original CBCT data. On the other hand, isotropic down-sampling produces 976 patches that occupy only 242 MB, which is around 5 times smaller than the previous memory consumption. Patches from the original CBCT also require 3 to 5 times higher evaluation time compared to the isotropic volume patches. Besides, a better evaluation result is obtained for patches with isotropic spacing. This is because the model is trained with patches converted to isotropic spacing. After observing the predictions, we find that the network shows comparatively poor performance near the metal foramen where the canal follows a curved path. One reason for this is that in some cases the given control points do not cover the full curved path. It is expected to have better results once more patches of this region will be added to the training dataset. Finally, We train our model with traditional region-based fixed diameter tubular-shaped ground truth to compare performance against our centerline ground truth. The result in Table 3 clearly shows that region-based ground truth is not required to trace mandibular canal routes. Rather, the centerline-based model predicts fewer false-positive canal voxels. A lower MCD-PG (0.65 mm against 0.86 mm for region-based) is such an indication.

## Conclusion

In this study, we propose a fully automatic deep learning-based framework to segment the mandibular canal. Results show that even the centerlines of the mandibular canals, used as model ground truths, can predict the canal pathway effectively. We consider this study as an opening for developing planning software for dental implement surgeries. Future studies include reduced processing time and less memory burden. A faster data processing is always expected by the end-users. Our initial plan is to extract the region-of-interest (ROI) first that will have only two rami along with a horseshoe-shaped body. This area is enough for canal extraction. The memory requirement for 3D CNN architectures is still a barrier to make it runnable on some mobile devices. We plan to extend the dataset with more diversity taking CBCT data from different scanners and different ethnicities. Finally, this study also encourages us to apply deep learning-based approaches to trace down other nerve-like structures.